\pdfoutput=1

\documentclass[11pt]{article}

\usepackage{emnlp2021}

\usepackage{times}
\usepackage{latexsym}

\usepackage[T1]{fontenc}

\usepackage[utf8]{inputenc}

\usepackage{microtype}
\usepackage{graphicx}
\usepackage[ruled,vlined]{algorithm2e}
\usepackage{amsmath}
\usepackage{multirow}
\usepackage{booktabs}
\usepackage{pgfplots}

\DeclareMathAlphabet\mathbfcal{OMS}{cmsy}{b}{n}
\makeatletter
\def\@fnsymbol#1{\ensuremath{\ifcase#1\or \dagger\or *\or \ddagger\or
   \mathsection\or \mathparagraph\or \|\or **\or \dagger\dagger
   \or \ddagger\ddagger \else\@ctrerr\fi}}

\title{Distilling the Knowledge of Large-scale Generative Models \\ into Retrieval Models for Efficient Open-domain Conversation}

\author{Beomsu Kim\thanks{\; Equal contribution}\;,\, Seokjun Seo\footnotemark[1]\;,\, Seungju Han\footnotemark[1]\;,\, Enkhbayar Erdenee\footnotemark[1]\;,\, Buru Chang\thanks{\; Corresponding author} \\
  Hyperconnect \\
  \{\small\texttt{beomsu.kim,seokjun.seo,seungju.han,enkhbayar.erdenee,buru.chang\}@hpcnt.com} \\
}

\begin{document}
\maketitle

\setlength{\abovedisplayskip}{3pt}

\begin{abstract}\label{sec:0_abstract}


Despite the remarkable performance of large-scale generative models in open-domain conversation, they are known to be less practical for building real-time conversation systems due to high latency.
On the other hand, retrieval models could return responses with much lower latency but show inferior performance to the large-scale generative models since the conversation quality is bounded by the pre-defined response set.
To take advantage of both approaches, we propose a new training method called G2R (Generative-to-Retrieval distillation) that preserves the efficiency of a retrieval model while leveraging the conversational ability of a large-scale generative model by infusing the knowledge of the generative model into the retrieval model.
G2R consists of two distinct techniques of distillation:
the data-level G2R augments the dialogue dataset with additional responses generated by the large-scale generative model, and the model-level G2R transfers the response quality score assessed by the generative model to the score of the retrieval model by the knowledge distillation loss.
Through extensive experiments including human evaluation, we demonstrate that our retrieval-based conversation system trained with G2R shows a substantially improved performance compared to the baseline retrieval model while showing significantly lower inference latency than the large-scale generative models.
 
\end{abstract}


\section{Introduction}\label{sec:1_introduction}
Recently, generative models have shown great success in open-domain conversation along with the development of large-scale language models, yielding fluent and informative responses \citep{roller2021recipes,adiwardana2020towards,NEURIPS2020_1457c0d6}.
However, generative models suffer from the challenges of latency and computational resources for building real-time conversation systems due to auto-regressive decoding for response generation and a large GPU memory footprint.



\begin{figure}[t]
\centering
\includegraphics[width=\columnwidth]{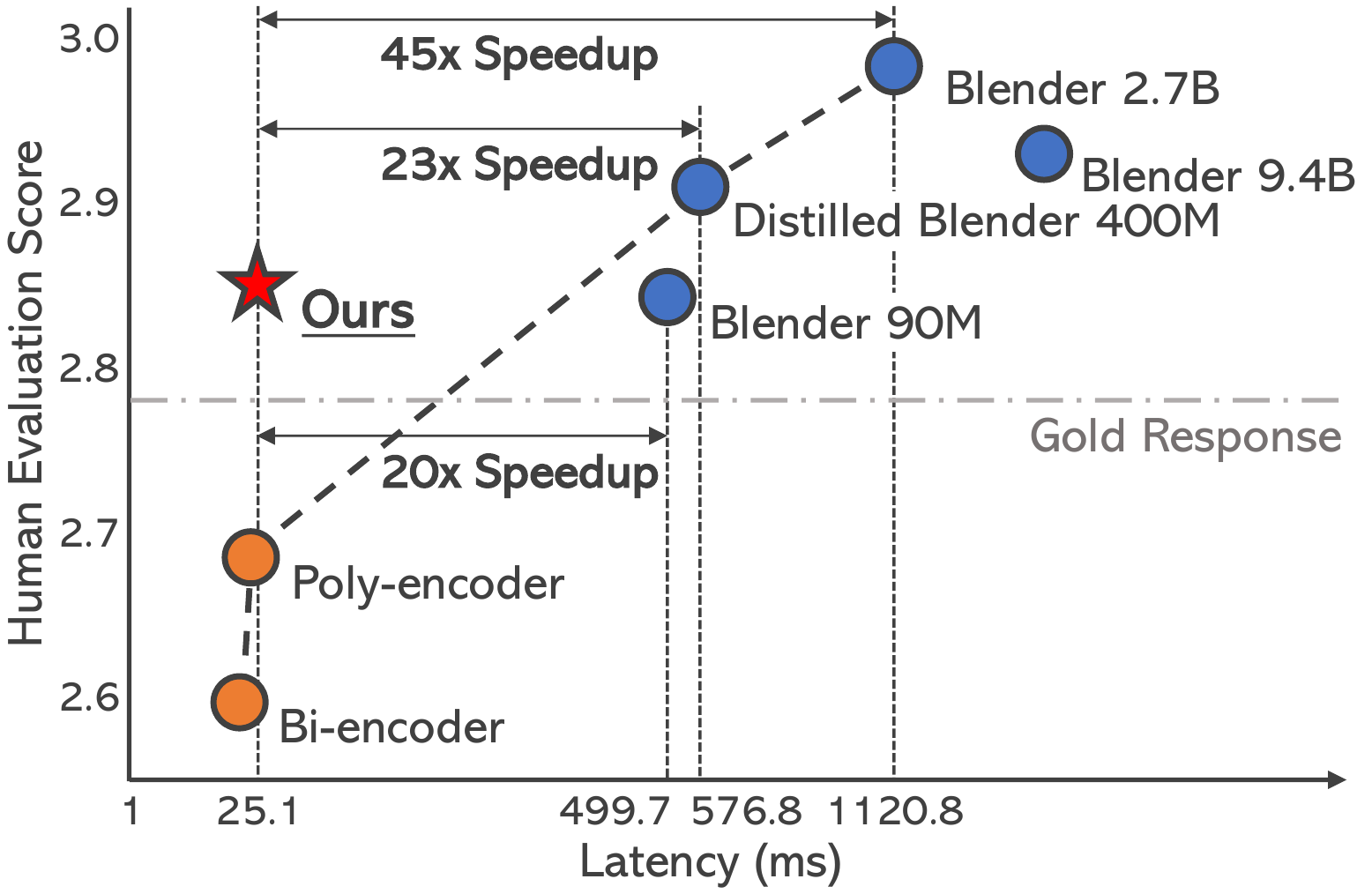}
\caption{
  Latency vs. Human evaluation score plot for open-domain conversation models.
  Blue circle represents generative models, orange circle represents retrieval models, and red star represents our model.
  Our model achieves a "sweet-spot" among various models, showing a much better human evaluation score than retrieval models and demonstrating much lower latency than generative models.
} 
\vspace*{-1.5em}
\label{fig:fig_2_human_evaluation}
\end{figure}

Meanwhile, retrieval models such as Bi-encoder and Poly-encoder \citep{humeau2019poly} is able to build efficient open-domain conversation systems by pre-defining the response set and searching the most relevant response to the given context from the response set.
In addition, a Bi-encoder dramatically reduces the latency when adopting efficient Maximum Inner Product Search (MIPS) libraries, such as FAISS \citep{johnson2019billion} and ScaNN \citep{avq_2020}.
Despite the outstanding efficiency, retrieval models have shown some lack of conversational ability compared to generative models. 
Retrieval models are known to return an erroneous response when the pre-defined response set does not contain the proper response to the given context, while generative models deal with these cases more flexibly \citep{weston2018retrieve}.
Exemplar-based generative models \citep{weston2018retrieve, wu2019response, gupta2021controlling} try to mitigate this problem by combining the advantages of the two approaches, whereas the inherent inefficiency of the generative models remains since exemplar-based generative models employ a generative model for response generation.

To make an efficient yet fluent open-domain conversation system, which is mandatory for real-world applications, we propose a novel training method for retrieval models called \textbf{\textit{Generative-to-Retrieval distillation}} (G2R).
G2R enables retrieval models to leverage the knowledge of large-scale generative models in both data-level and model-level.
First, \textbf{\textit{data-level G2R}} augments the original dialogue dataset with the responses produced by a large-scale generative model using contexts in the original dialogue dataset.
Then, the produced responses are also added to the pre-defined response set.
The augmented dialogue dataset and response set are utilized for training a retrieval model at the training phase and for returning responses at the inference phase, respectively.
Although data-level G2R enables retrieval model to utilize high-quality responses generated by the large-scale generative model, it does not transfer the fine-grained knowledge from the generative model about the quality of individual responses.
\textbf{\textit{Model-level G2R}} resolves this limitation by transferring the response quality scores assessed by the large-scale teacher generative model into the scores of the student retrieval model.
This method induces the retrieval model to select a better response in terms of the response quality.

We empirically demonstrate that a retrieval-based conversation system, which consists of the G2R-applied retrieval model and a MIPS library, shows a substantial conversational ability while showing fast inference speed, as shown in Figure \ref{fig:fig_2_human_evaluation}.
For instance, our retrieval-based conversation system shows about a 20x speedup compared to the Blender model (90M parameters) while exhibiting a comparable human evaluation result on conversational ability. 

\section{Method}\label{sec:2_method}
\subsection{Preliminaries}\label{seubsec:2_1_reliminaries}
\textbf{Retrieval models for Open-domain Conversation.}
Let $D=\{(c_i,r_i) \mid 1 \le i \le n\}$ denote the dialogue dataset that contains $n$ context-response pairs, where $c_i$ and $r_i$ are a context and its corresponding gold response of the $i$-th example, respectively.
At the training phase, retrieval models are trained to maximize the score of the gold response $r_i$ for the given context $c_i$ compared to the scores of negative responses.
At the inference phase, retrieval models return the response with the highest score for the given context $c$ from the pre-defined response set $R=\{r_i \mid 1 \le i \le n\}$ constructed from the dialogue dataset $D$.
\newline
\textbf{Knowledge Distillation.}
Knowledge Distillation \cite{hinton2015distilling} transfers the knowledge of the teacher model into the student model by adding a loss that matches the logits of the student model $z_s$ with the logits of the teacher model $z_t$.
For classification task with $l$ classes, the knowledge distillation loss is defined by the cross-entropy between the softened output probability of the student model and the teacher model:
\begin{multline}
\small
    \mathcal{L_{KD}} = -\sum_{x \in X} \sum_{i=1}^l p_t(y_i|x) \log p_s(y_i|x) \\
      = -\sum_{x \in X}\sum_{i=1}^l \left[ \frac{\exp(z_t(x, y_i)/T)}{\sum_j{\exp(z_t(x, y_j)/T)}} \times \right. \\
     \left. \log \frac{\exp(z_s(x, y_i)/T)}{\sum_j{\exp(z_s(x, y_j)}/T)} \right],
  \label{eq:knowledge_distillation_loss}
\end{multline}
where $p(y|x)$ and $z(x, y)$ are the softened probability and logit value of the models for the input $x$ and class $y$, respectively, and $T$ is a temperature parameter for smoothing the logit values.

\subsection{Retrieval-based Conversation System}\label{sec:2_2_TwoStagePipeline}

Our goal is to create an efficient open-domain conversation system based on the retrieval model.
However, naively utilizing the retrieval model can lead to the low efficiency when the size of the response set $R$ is large since the retrieval model has to calculate scores for all response candidates.
To this end, we adopt the Bi-encoder \citep{humeau2019poly} model with an efficient MIPS library to select proper responses efficiently without calculating a score for all response candidates.
Bi-encoder encodes a context $c$ and response $r$ into the fixed-length embedding respectively with Transformer architecture \citep{vaswani2017attention}, and defines the relevance score between $c$ and $r$ as the dot-product of two embeddings.
Therefore, an efficient MIPS library, FAISS \citep{johnson2019billion} for our case, can be utilized for speeding up the search process.

\subsection{Data-level G2R}\label{subsec:2_3_data_level_g2r}
\begin{figure*}[t]
\centering
\includegraphics[width=\textwidth]{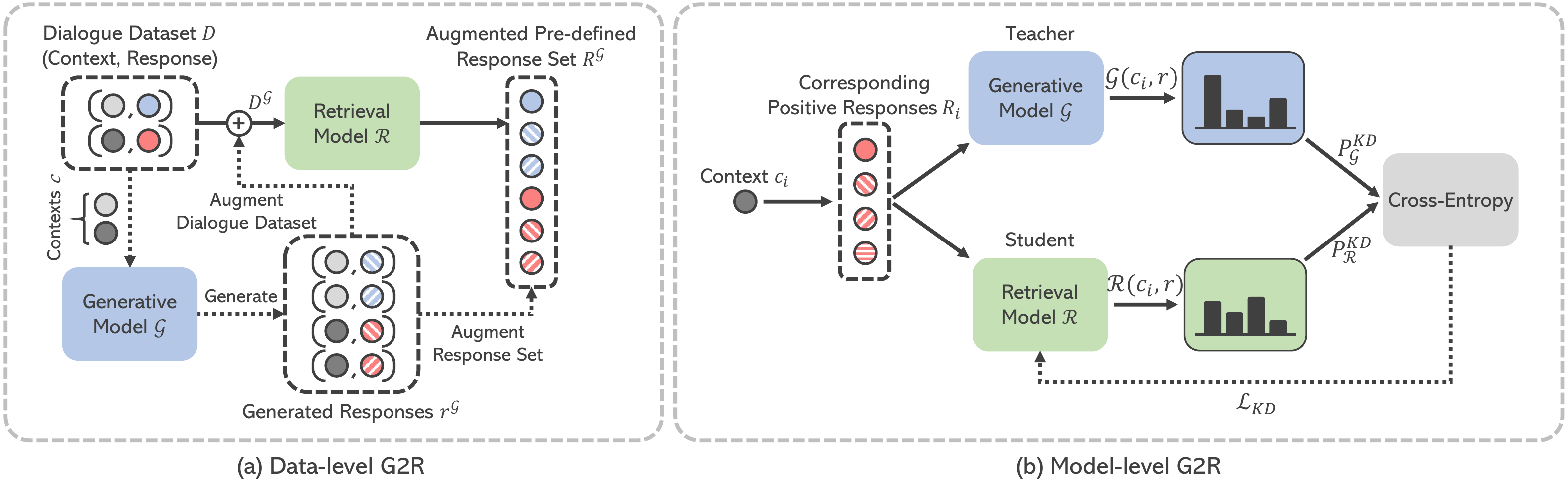}
\caption{Illustration of our proposed training method G2R.}
\vspace*{-1em}
\label{fig:fig_1_architecture}
\end{figure*}

It is well-known that utilizing an additional high-quality dialogue dataset is helpful for improving the performance of the retrieval model, as shown in \citet{zhang2020dialogue}.
Moreover, enriching the pre-defined response set $R$ with more diverse responses can help the retrieval model to respond appropriately to a variety of input contexts since it widens the opportunity to select an appropriate response.
However, it is highly labor-intensive and costly to acquire such high-quality dialogue datasets or responses through human-in-the-loop annotation such as in \citet{zhang2018personalizing} or \citet{smith2020can}.

Meanwhile, previous studies \cite{adiwardana2020towards, roller2021recipes,NEURIPS2020_1457c0d6} show that well-tuned large-scale generative models are able to achieve near-human conversational ability.
From these observations, we are motivated to leverage the generation result of large-scale generative models to extend the response set as well as the dialogue dataset for training a retrieval model, as shown in Figure \ref{fig:fig_1_architecture}(a).

For each context $c_i$ in the dialogue dataset $D$, a large-scale generative model $\mathcal{G}$ generates $m$ responses, $\{r^{\mathcal{G}}_{i,j} \mid 1 \le j \le m \}$.
Considering the generated responses as a gold response of the given context $c_i$, they are added to the dialogue dataset $D$ and the pre-defined response set $R$ as follows:
$D^{\mathcal{G}}= D \cup \{(c_i, r^\mathcal{G}_{i,j}) \mid 1 \le i \le n, 1 \le j \le m\}$ and $R^{\mathcal{G}} = R \cup \{r^\mathcal{G}_{i,j} \mid 1 \le i \le n, 1 \le j \le m\}$.
$D^{\mathcal{G}}$ and $R^{\mathcal{G}}$ denote the augmented dialogue dataset and response set, respectively.

After the augmentation, a retrieval model $\mathcal{R}$ is trained by minimizing the cross-entropy loss $\mathcal{L}_{CE}$ which maximizes the probability of selecting the ground-truth response $r$ among the set of randomly sampled negative responses $R^-$:
\begin{equation}
\mathcal{L}_{CE} = -\sum_{(c,r) \in \mathcal{D}^\mathcal{G}} \log\frac{\exp(\mathcal{R}(c, r))}{\sum_{r^- \in \{r\} \cup R^-} \exp(\mathcal{R}(c, r^-))},
\label{eq:data_distillation_loss}
\end{equation}
where $\mathcal{R}(c, r)$ is the score computed by the retrieval model $\mathcal{R}$ for the given context $c$ and response $r$.
Note that $R^-$ is created differently for every iteration by randomly sampling responses from $R^\mathcal{G}$ without replacement.

We employ the largest open-domain conversation model available, Blender 9.4B \citep{roller2021recipes}, as the large-scale generative model $\mathcal{G}$.
We apply top-k sampling \citep{fan2018hierarchical} for the diversity of responses since beam search tends to generate similar responses within the same context \citep{adiwardana2020towards}.
In addition, we sample responses multiple times with different minimum length constraints to diversify the specificity and length of generated responses.

\subsection{Model-level G2R}\label{subsec:2_4_model_level_g2r}

While data-level G2R provides additional high-quality dialogue data and diverse responses, it does not transfer the fine-grained knowledge about the quality of the individual responses from the large-scale generative model $\mathcal{G}$.
Model-level G2R is designed to address this problem by transferring the individual response-level quality score, assessed by the large-scale teacher generative model $\mathcal{G}$, into the student retrieval model $\mathcal{R}$.
We first define the quality score of the response from the perspective of the teacher generative model $\mathcal{G}$, denoted as $\mathcal{G}(c, r)$.
Then, the student retrieval model is trained to match the score $\mathcal{R}(c, r)$ of the student retrieval model with the score $\mathcal{G}(c, r)$ of the teacher generative model, similar to the conventional knowledge distillation technique \citep{hinton2015distilling}.
Overall process of knowledge distillation is depicted in Figure \ref{fig:fig_1_architecture}(b).

We define the generator score $\mathcal{G}(c, r)$ as the log-likelihood normalized by the length of response:

\begin{equation}
    \mathcal{G}(c, r) = (\log P_{\mathcal{G}}(r|c)) / |r|,
    \label{eq:log_likelihood_score}
\end{equation}
where $P_\mathcal{G}(r|c)$ is the probability of the response $r$ for the given context $c$ of the generative model $\mathcal{G}$ and $|r|$ is the number of tokens in the response $r$.
Log-likelihood is normalized with the length of response to mitigate the problem of preferring shorter responses  \citep{murray2018correcting}.

We can derive the distillation loss $\mathcal{L}_{KD}$ by regarding the generator quality score $\mathcal{G}(c, r)$ and retriever score $\mathcal{R}(c, r)$ as the logits of teacher and student model, respectively.
Eq. \ref{eq:knowledge_distillation_loss} then turns into:

\begin{equation}
  \begin{gathered}
    P_\mathcal{G}^{KD}(c_i, r) = \frac{\exp(\mathcal{G}(c_i, r) / T)}{\sum_{r' \in R_i \cup R^-} \exp(\mathcal{G}(c_i, r') / T)}, \\
    P_\mathcal{R}^{KD}(c_i, r) = \frac{\exp(\mathcal{R}(c_i, r) / T)}{\sum_{r' \in R_i \cup R^-} \exp(\mathcal{R}(c_i, r') / T)}, \\
    \mathcal{L}_{KD} = -\sum_{i=1}^n \sum_{r \in R_i \cup R^-} P_\mathcal{G}^{KD}(c_i, r) \log P_\mathcal{R}^{KD}(c_i, r), 
  \end{gathered}
\end{equation}
where $R_i = \{r_i, r_{i,1}^\mathcal{G}, \cdots, r_{i,m}^\mathcal{G}\}$ is a set of positive responses correspond to the context $c_i$ in $D^\mathcal{G}$.
Since calculating the generator quality score $\mathcal{G}(c_i, r^-)$ for negative responses requires heavy extra computation, we simplify the calculation by approximating $P_\mathcal{G}^{KD}(c_i, r^-) \approx 0$, $\exp (\mathcal{G}(c_i, r^-) / T) \approx 0$ for randomly sampled negative responses $r^- \in R^-$.

Our final loss for the model-level G2R is a sum of original cross-entropy loss in Equation \ref{eq:data_distillation_loss} and the knowledge distillation loss where hyperparameter $\alpha$ controls the weights of each term:

\begin{equation}
\mathcal{L} = \alpha \mathcal{L}_{CE} + (1 - \alpha) \mathcal{L}_{KD}.
\label{eq:model_distillation_loss}
\end{equation}

\section{Experiments}\label{sec:3_experiments}

\subsection{Dataset}
We conduct experiments on the open-domain conversation datasets which consist of Blended Skill Talk \citep{smith2020can}, ConvAI2 \citep{zhang2018personalizing}, Empathetic Dialogues \citep{rashkin2019towards} and Wizard of Wikipedia \citep{dinan2018wizard}.
Following \citet{roller2021recipes}, all four datasets are used together for the following experiments, and we refer to the merged dataset as BST+.
We follow the method of splitting train, validation, and test set from \citet{smith2020can}.

\subsection{Metrics}

\textbf{Human Evaluation.}
We conduct a human evaluation to assess the quality of model responses.
Human evaluation is carried out on 200 examples randomly sampled from the BST+ test dataset.
Human judges are asked to evaluate the quality of the generated response with two criteria on a 0-2 scale: (i) \textit{Appropriateness (Appr.)} for evaluating whether the generated response is fluent, logical, and appropriate to its given context, and (ii) \textit{Informativeness (Info.)} for evaluating whether the generated response has meaningful information relevant to its given context.
Each example is rated by at least three unique human judges, and all the human evaluation is performed via Amazon Mechanical Turk.
\newline
\textbf{Automated Metrics.}
We also report various kinds of automated metrics.
\textit{MaUdE} \cite{sinha2020learning} is an unreferenced dialogue response evaluation metric calculated by the model that is trained to score positive responses as 1 while scoring syntactically and semantically negative responses as 0, using the ConvAI2 dataset.
Since \textit{MaUdE} shows a high correlation with human judgments on fluency and interestingness of responses, we use \textit{MaUdE} as a proxy metric for evaluating the overall quality of responses produced by each model.
For measuring the lexical diversity of generated responses we utilize \textit{Dist-2} and \textit{Dist-3} \citep{li2016diversity}, where \textit{Dist-n} is a ratio of distinct n-grams to the total number of n-grams in all the responses generated by each model.
\textit{Length}, the average number of tokens in generated responses, is reported for reference.
Last but not least, we measure and report the \textit{Latency} for generating a response for a single input context to verify the efficiency of the model.
Although we report the latency measured on the GPU-enabled environment, the latency measured by using only the CPU is reported in the supplementary material.

\begin{table*}[t]
\centering
\footnotesize
\begin{tabular}{cccccccccc}
\toprule
\multicolumn{1}{c}{\multirow{2}{*}{Models}} & \multicolumn{3}{c}{Human Evaluation}                                       & \multicolumn{4}{c}{Automated Metrics}                                                  & \multicolumn{2}{c}{\multirow{2}{*}{\shortstack{\\Latency \\ (ms)}} \multirow{2}{*}{\shortstack{\\Latency \\ (Speedup)}}} \\ \cmidrule(lr){2-4} \cmidrule(lr){5-8}
\multicolumn{1}{c}{}                        & \multicolumn{1}{c}{Sum} & \multicolumn{1}{c}{Appr.} & \multicolumn{1}{c}{Info.}  & \multicolumn{1}{c}{MaUdE} & \multicolumn{1}{c}{Dist-2} & \multicolumn{1}{c}{Dist-3} & \multicolumn{1}{c}{Length} & \multicolumn{2}{c}{}                         \\
\midrule
Blender 90M & 2.843  & 1.429 & 1.414 & 0.8582 & 0.4799 & 0.6887 & 18.31 & 499.7 & \hspace{1ex}1.00x                                              \\
Blender 2.7B & 2.983 & 1.510 & 1.473 & 0.8826 & 0.5288 & 0.7261 & 19.05 & 1120.8 &\hspace{1ex}0.45x                                     \\
Blender 9.4B & 2.930 & 1.472 & 1.458 & 0.8763 & 0.5246 & 0.7285 & 18.87 & 1438.6 & \hspace{1ex}0.35x                                              \\ 
Distilled Blender & 2.910 & 1.474 & 1.436 & 0.8715 & 0.4821 & 0.6815 &  19.19 & 576.8 & \hspace{1ex}0.87x                                           \\ 
RetNRef & 2.771 & 1.404 & 1.368 & 0.8555 & 0.7773 & 0.9541 & 12.34 & 382.4 &\hspace{1ex}1.31x                                     \\ 
Bi-encoder & 2.597 & 1.288 & 1.309 & 0.8944 & 0.8191 & 0.9712 & 14.85 & 18.6 &\hspace{0.1ex}26.87x                                      \\
Poly-encoder & 2.686 & 1.340 & 1.346 & 0.8645 & 0.8269 & 0.9692 & 15.30 & 24.8 &\hspace{0.1ex}20.15x                                        \\ 
Bi-encoder (w/ FAISS) & 2.596 & 1.259 & 1.337 & 0.9046 & 0.8316 & 0.9735 & 15.22 & 25.7 & 19.44x    \\
\midrule
G2R-D (w/o FAISS)  & 2.779 & 1.380 & 1.399 & 0.8518 & 0.7242 & 0.9302 & 20.06 & 39.7 & 12.59x \\
G2R-D  & 2.759 & 1.398 & 1.361 & 0.8443 & 0.7456 & 0.9395 & 19.93 & 25.3 & 19.75x \\
G2R-DM & 2.856 & 1.447 & 1.410 & 0.8695 & 0.7266 & 0.9393 & 17.48 & 25.1 & 19.91x \\
\midrule
Human Response & 2.788 & 1.418 & 1.369 & 0.9146 & 0.8271 & 0.9742 & 14.22 & - & - \\
\bottomrule

\end{tabular}
\caption{Human evaluation results and automated metrics of the baseline models and our G2R models. \textit{Latency (Speedup)} column denotes the relative speedup of each model compared to the latency of \textit{Blender 90M}.}
\vspace*{-2mm}
\label{tab:main_result}
\end{table*}

\subsection{Models and Baselines} 

\textbf{Blender.}
Blender, the state-of-the-art model in open-domain conversation task, is adtoped with different number of parameters: \textit{Blender 90M}, \textit{Blender 2.7B}, and \textit{Blender 9.4B}.
For response generation, we follow the decoding hyperparameters suggested in the original work \citep{roller2021recipes}.
\newline
\textbf{Distilled Blender.}
A small Blender model distilled from a larger generative model is employed to compare our result with a generative model that also utilizes the knowledge distillation technique.
We use 400M parameters Blender model distilled from \textit{Blender 2.7B} with TinyBERT style distillation \citep{jiao2020tinybert}, denoted as \textit{Distilled Blender}.
\newline
\textbf{Bi-encoder \& Poly-encoder.}
\textit{Bi-encoder} and \textit{Poly-encoder} with 256M parameters \citep{humeau2019poly}, pre-trained with the Pushshift Reddit comment dataset \citep{baumgartner2020pushshift} and fine-tuned on the BST+ dataset, are the baselines for retrieval models.
The \textit{Bi-encoder} model integrated with MIPS library, as described in Section \ref{sec:2_2_TwoStagePipeline}, is denoted as \textit{Bi-encoder (w/ FAISS)}.
\newline
\textbf{RetNRef.}
\textit{RetNRef} \citep{weston2018retrieve} is an exemplar-based generative model which incorporates the response of retrieval models into the input of the generative model.
Contrary to G2R, \textit{RetNRef} exploits the retrieval model to make the generative model better, while G2R exploits the knowledge of the generative model to make the retrieval model better.
We use the dialogue retrieval model described in \citet{roller2021recipes} trained with the $\alpha$-blending technique.
\newline
\textbf{Human Response.}
\textit{Human response} refers to the ground-truth label annotated in the BST+ dataset.
\newline
\textbf{G2R.}
Our system is built upon the retrieval-based conversation system described in Section \ref{sec:2_2_TwoStagePipeline}, where the Bi-encoder $\mathcal{R}$ is trained with our proposed G2R using \textit{Blender 9.4B} as the teacher generative model $\mathcal{G}$.
\textit{G2R-DM} denotes our model trained with the data-level G2R and the model-level G2R.
For a comprehensive analysis, two variants are adopted:
\textit{G2R-D} is trained with the data-level G2R only, and \textit{G2R-D (w/o FAISS)} further removes the use of the MIPS library, FAISS, from \textit{G2R-D}.

\subsection{Implementation Details}
We provide the details on our implementation and the hyperparameter values in the supplementary material.
For reproducibility, we release the augmented dialogue dataset and the implementation of G2R models.\footnote{\url{https://github.com/hyperconnect/g2r}}
\section{Experimental Results}\label{sec:5_experimental_results}

\subsection{Result Analysis}

We present the human evaluation result and the automated metrics in Table \ref{tab:main_result}.
Overall, our system trained with G2R achieves a "sweet-spot" between conversational ability and efficiency.
Our system maintains the low latency of \textit{Bi-encoder (w/ FAISS)} while boosting up the human evaluation results significantly, achieving comparable or better human evaluation scores than the \textit{Blender 90M} and human responses, respectively.

Taking a closer look, the Blender generative models and the distilled variant show high human evaluation metric while showing relatively large latency along with the lack of diversity, as shown in the Dist-2 and Dist-3 scores.
Retrieval baselines (\textit{Bi-encoder} and \textit{Poly-encoder}) show an opposite trend, exhibiting much lower latency and relatively higher response diversity but showing relatively lower conversational ability in terms of human evaluation metric.
Unlike human evaluation results, the MaUdE scores of the \textit{Bi-encoder} and the \textit{Poly-Encoder} are unexpectedly high.
However, we suspect this is because the MaUdE metric is trained on the ConvAI2 dataset, which is a subset of the BST+ dataset, and with a similar training objective of these retrieval models as described in Section \ref{sec:3_experiments}.

G2R-based models achieve far better human evaluation results compared to their original model, \textit{Bi-encoder (w/ FAISS)}.
Applying data-level G2R only (\textit{G2R-D}) significantly boosts the performance, making the model perform comparable to gold human response in terms of human evaluation.
Using data-level G2R enlarges the number of responses in the pre-defined response set $R^\mathcal{G}$ more than ten times, therefore using Bi-encoder without FAISS (\textit{G2R-D (w/o FAISS)}) leads to increased latency.
Although using FAISS induces a latency overhead for the case where the size of the response set is small (\textit{Bi-encoder (w/ FAISS)}), using FAISS in a larger response set as in \textit{G2R-D} enables us to maintain the low latency, while having a slight degradation of response qualities compared to the version without FAISS.


Further application of model-level G2R additionally boosts the performance of the retrieval model.
\textit{G2R-DM} shows a higher human evaluation score and MaUdE score than \textit{G2R-D}, and exhibits a comparable human evaluation score to the \textit{Blender 90M} model while running much faster.
While \textit{G2R-DM} shows a somewhat deficient human evaluation score compared to the bigger Blender generative models, it shows substantially lower latency (23.0x speedup over \textit{Distilled Blender}, 44.7x speedup over \textit{Blender 2.7B}).
In addition, \textit{G2R-DM} exhibits a much higher response diversity compared to the Blender generative models.
The \textit{RetNRef} model shows worse performance and delivers much higher latency compared to our \textit{G2R-DM} model. 

\subsection{Statistics of the Responses augmented by the Data-level G2R}
\begin{table}[t]
\centering
\footnotesize
\begin{tabular}{c|cc|c}
\toprule
Statistics                  & $R$ & $R^\mathcal{G}$ & Ratio \\
\midrule
\# of Responses     & 279,090 & 3,070,074 & 11.0x \\
Average length    & 14.85  & 18.78 & 1.26x \\
\# of Unique Tokens  & 56,862  & 210,538 & 3.70x \\
\# of Unique bi-grams & 655,948 & 2,710,155 & 4.13x \\
\# of Unique tri-grams & 1,738,189 & 10,654,181 & 6.13x \\
\bottomrule
\end{tabular}
\caption{Comparison of the statistics of the original response set $R$ and the response set $R^\mathcal{G}$ augmented by data-level G2R. We also report the ratio of the statistics of the augmented dataset to those of the original dataset.}
\vspace*{-1.5em}
\label{tab:data_distillation_response_stats}
\end{table}


\begin{table*}[t]
\footnotesize
\centering
\begin{tabular}{@{}cccccccccc@{}}
\toprule
\multicolumn{1}{c}{\multirow{2}{*}{\shortstack{\\ Train \\ $\mathcal{R}$ with}}} & \multicolumn{1}{c}{\multirow{2}{*}{\shortstack{\\Response \\ Set}}}   & \multicolumn{3}{c}{Human Evaluation} & \multicolumn{5}{c}{Automated Metrics} \\ \cmidrule(lr){3-5} \cmidrule(lr){6-10}
 &  & Sum & Appr. & Info. & Dist-2            & Dist-3         & Length & Hits@1/K & Hits@5/K \\
\midrule
\multirow{2}{*}{$D$} & $R$ & 2.596 & 1.259 & 1.337 & 0.8336 & 0.9777 & 15.66 & \multirow{2}{*}{0.7537} & \multirow{2}{*}{0.9363} \\
 & $R^\mathcal{G}$ & 2.620 & 1.300 & 1.320 & 0.7660 & 0.9498 & 17.14 \\
 \midrule
\multirow{2}{*}{$D^\mathcal{G}$} & $R$ & 2.739 & 1.377 & 1.361 & 0.8144 & 0.9687 & 16.20 & \multirow{2}{*}{0.8052} & \multirow{2}{*}{0.9570} \\
 & $R^\mathcal{G}$ & 2.770 & 1.403 & 1.368 & 0.7456 & 0.9395 & 19.93 \\
 \midrule
 \midrule
 \textit{$D^\mathcal{R}$} & $R$ & 2.591 & 1.296 & 1.295 & 0.8253 & 0.9669 & 14.54 & 0.7594 & 0.9323 \\
\bottomrule
\end{tabular}
\caption{Human evaluation and automated metric results of the ablation models for data-level G2R. 
Note that $D^\mathcal{R}$ is inspired from \citet{zhu2020data}, and is not the G2R method.}
\vspace*{-3mm}
\label{tab:ablation_data_distillation}
\end{table*}

Table \ref{tab:data_distillation_response_stats} shows the basic statistics of the original response set $R$ and the response set $R^\mathcal{G}$ created by data-level G2R.
After applying the data-level G2R, $R^\mathcal{G}$ has roughly 11 times more candidates compared to the original response set $R$.
To verify if responses in the new response set $R^\mathcal{G}$ show more diversity, we count the number of unique tokens and bi-gram/tri-grams appearing in each response set.
The augmented response set has much more unique tokens and bi-gram/tri-grams than the original response set, implying that it covers more diverse topics, entities and shows more diversity in terms of phrases and expressions.

\subsection{Ablation Studies}
\textbf{Breakdown analysis of Data-level G2R. }  We conduct an ablation study to analyze in detail how the performance of the model changes depending on how we use responses generated in the data-level G2R method.
In data-level G2R, generated responses are utilized in two ways: for augmenting the training dialogue dataset $D^{\mathcal{G}}$ of the retrieval model $\mathcal{R}$, and for building the augmented response set $R^\mathcal{G}$.
We separate these two utilization methods and evaluate models that use only each method.

Table \ref{tab:ablation_data_distillation} shows the evaluation results of these ablation models.
Along with the human evaluation metrics and automated metrics, we also report \textit{Hits@1/K} and \textit{Hits@5/K} \citep{roller2021recipes} of trained Bi-encoder model on the BST+ test set, which are widely adopted to evaluate the performance of retrieval models.
As shown in the table, only utilizing one of the methods does not show better performance compared to the model that utilizes both methods.
Utilizing the generated responses for building $R^\mathcal{G}$ improves the appropriateness score of the model, which supports the hypothesis we have raised in Section \ref{sec:2_method} that using a diverse response set is helpful for the model to respond more appropriately.
The use of augmented dialogue $D^\mathcal{G}$ for training $\mathcal{R}$ is helpful for increasing a human evaluation score, for both appropriateness and informativeness metrics, meaning that the retrieval model learns to select relevant and rich responses that the generative model created.
In addition, training with augmented dialogue $D^\mathcal{G}$ considerably improves the Hits metric of the retrieval model.
Nonetheless, using both methods shows the best human evaluation performance among all ablation models, indicating that using new examples for both training a retrieval model and building a response set is crucial for inducing a good performance.
\newline
\textbf{Different Dialogue Augmentation Strategy. }
Here, we implement a simple baseline inspired by \citet{zhu2020data} and \citet{zhang2020dialogue}, which augments training dialogue by utilizing top-$m$ responses of a retrieval model that has already been trained.
In this experiment, we use the \textit{Bi-encoder} model for this augmentation process, and the augmented dialogue dataset generated by this method is denoted as $D^\mathcal{R}$.
Comparison of data-level G2R with this baseline will enable us to verify that our method with a large generative model produces better quality training dataset than simply using a retrieval model.

\begin{table}[t]
\centering
\footnotesize
\setlength\tabcolsep{3.0pt}
\begin{tabular}{ccccccc}
\toprule
\multicolumn{1}{c}{\multirow{2}{*}{$\mathcal{G}(c, r)$}} & \multicolumn{3}{c}{Human Evaluation}                                       & \multicolumn{3}{c}{Automated Metrics}  \\ 
\cmidrule(lr){2-4} \cmidrule(lr){5-7}
\multicolumn{1}{c}{} & \multicolumn{1}{c}{Sum} & \multicolumn{1}{c}{Appr.} & \multicolumn{1}{c}{Info.}  & \multicolumn{1}{c}{MaUdE} & \multicolumn{1}{c}{Dist-2} & \multicolumn{1}{c}{Dist-3} \\
\midrule
LL & 2.856 & 1.447 & 1.410 & 0.8695 & 0.7266 & 0.9393  \\
MI & 2.806 & 1.427 & 1.380 & 0.8737 & 0.7536 & 0.9468  \\
\bottomrule

\end{tabular}
\caption{Human evaluation results and automated metrics for model-level G2R models that use different score for defining generator quality score $\mathcal{G}(c, r)$.}
\vspace*{-1.5em}
\label{tab:ablation_model_distillation}
\end{table}

The result of this ablation study is reported in Table \ref{tab:ablation_data_distillation}.
As shown in the table, using $D^\mathcal{R}$ as the training dataset does not lead to a significant performance gain for all metrics, contrary to the case of using $D^\mathcal{G}$ which improves both human evaluation score and Hits metric. 
This result strongly indicates that utilizing a large-scale generative model for dialogue augmentation as in data-level G2R is a much more effective augmentation strategy than using retrieval models.
\newline
\textbf{Utilizing a Different Generator Quality Score for Model-level G2R. }
Although we employ the log-likelihood score (LL score) for defining the generator quality score $\mathcal{G}(c, r)$ in model-level G2R, there are other methods that can be utilized as well.
One example is a Mutual Information score (MI score) \citep{li2016diversity}, which is a point-wise mutual information between the given context $c$ and response $r$.
Details about calculating the MI score of response are described in the supplementary material.
MI score is known to assign lower values to generic responses while escalating the score of responses that are more specific to the given context, so we expect that using the MI score will produce more specific and diverse responses compared to the LL score while having a slightly higher risk of returning responses with inappropriate details with respect to input context.
Therefore, we evaluate the variant of model-level G2R that uses MI score as $\mathcal{G}(c, r)$ and compare the performance with the model that uses LL score.

The results are provided in Table \ref{tab:ablation_model_distillation}. 
Using MI score for the model-level G2R exhibits a slightly lower human evaluation score than using LL score, especially for the appropriateness score, implying that using the MI score may be less capable of producing an appropriate and accurate answer. 
However, in terms of automated metrics, the MI score shows a higher MaUdE score.
In addition, using MI score shows higher response diversity compared to LL score as expected, indicating that MI score could be employed for more diverse responses of the conversation system.

\subsection{Case Study} 
\begin{table*}[ht]
\centering
\footnotesize
\begin{tabular}{c|l|l|l|c}
\toprule
\multicolumn{5}{c}{\textbf{Input Context}}                                                  \\ \midrule
\multicolumn{5}{l}{\begin{tabular}[l]{@{}l@{}} A: Amazon is a great place to order books \\ B: Oh yeah that is a fact. \end{tabular}} \\ 
\midrule
\multicolumn{1}{c|}{\textbf{Model Name}}                 &  \multicolumn{4}{|c}{\textbf{Response}}                 \\ \midrule
Bi-encoder (w/ FAISS) & \multicolumn{4}{|l}{A: Comics is at the top of the list in interest}  \\
Blender 90M & \multicolumn{4}{|l}{A: Do you have any other hobbies that you like to do on the weekends? I like to read}  \\
Distilled Blender & \multicolumn{4}{|l}{A: What else do you like to do in your spare time? I like to go to the beach}  \\
Blender 2.7B & \multicolumn{4}{|l}{A: What kind of books do you buy on amazon? I like to read science fiction and fantasy} \\
Blender 9.4B & \multicolumn{4}{|l}{A: What kind of books do you like to read? I like romance novels and mystery novels} \\
\midrule
G2R-D & \multicolumn{4}{|l}{A: Do you ever order grocerys? I love amazon's selection of books and videos} \\
G2R-DM & \multicolumn{4}{|l}{A: What is your favourite book? Mine is "the cat in the hat" by dr seuss} \\
\bottomrule
\end{tabular}

\caption{Example responses in the BST+ test set example. Full dialogue context is shown in the supplementary material.}
\vspace*{-1em}
\label{tab:qualitative_results}
\end{table*}
Table~\ref{tab:qualitative_results} provides an example of responses returned by the baseline models and our G2R models.
In this example, \textit{Bi-encoder (w/ FAISS)} returns the irrelevant response to the given context. 
Blender models' responses are logically appropriate, however, they just simply change the topic (\textit{Blender 90M}, \textit{Distilled Blender}) or relatively lack of specific details (\textit{Blender 2.7B}, \textit{Blender 9.4B}).
\textit{G2R-D} tries to respond with detail, but the response contains a somewhat irrelevant phrase about groceries.
In contrast, \textit{G2R-DM} respond appropriately along with specific details talking about a particular book title.
We provide additional response examples in the supplementary material.

\section{Related Work}\label{sec:2_related_work}

\subsection{Open-domain Conversation}
The task of open-domain conversation has been studied based on retrieval models, generation models, or using both.
While retrieval models~\citep{wang2013dataset,ji2014information,wang2015syntax,yan2016learning,wu2017sequential,zhou2018multi,tao2019one,humeau2019poly} search a response relevant to a given context from a pre-defined response set, generative models~\citep{shang2015neural,vinyals2015neural,li2020don,holtzman2019curious,welleck2019neural,roller2021recipes} produce a response based on the given context with auto-regressive decoding.
It is known that the retrieval and generative models have advantages in the efficiency of inference and quality of generated responses, respectively.
To take both advantages, several exemplar-based generative models~\citep{guu2018generating,wu2019response,weston2018retrieve,cai2019retrieval,gupta2021controlling} have recently been proposed by combining the retrieval and generative models.
The main difference between our proposed training method and the exemplar-based generative models is that exemplar-based generative models provide the knowledge of retrieval models to generative models, while our proposed training method transfers the knowledge of generative models to retrieval models to focus on the efficiency of open-domain conversation systems.

\subsection{Knowledge Transfer from Large Models}
Transferring the knowledge from larger-scale teacher neural networks into smaller-scale student neural networks has been implemented to improve the performance of the student model, including data augmentation and knowledge distillation.
In the data augmentation perspective, several works \citep{schick2021generating, chang2021jointly, kumar2020data, yang2020g} utilize the generation result of pre-trained language models as a labeled example for text classification tasks.
\citet{lin2020world} utilize the inference result of the retrieval model and the generative model as a semi-negative dataset for training a student retrieval model.
Meanwhile, Knowledge distillation \citep{hinton2015distilling} transfers the knowledge of the teacher model into the student model by matching the student logits with softened teacher logits.
Knowledge distillation especially designed for specific tasks or model architectures exists, such as sequence generation task \citep{kim2016sequence, lin2020autoregressive}, retrieval models \citep{lu2020twinbert, vakili2020distilling} and for transformer architectures \citep{jiao2020tinybert, wang2020minilm, sun2020mobilebert}.

The most related work to our paper is Dialogue Distillation \citep{zhang2020dialogue}, which also proposes a data-level and model-level distillation for open-domain conversation models.
Our research differs from this work in three ways.
First, Dialogue Distillation requires additional unpaired text corpus, which could be hard to be obtained in certain circumstances.
We instead focus on utilizing the knowledge of large-scale generative models for augmenting additional data.
In addition, Dialogue Distillation does not enrich the pre-defined response set, which is crucial for improving the performance of the retrieval models, as shown in our experiments.
Last but not least, while Dialogue Distillation only considers the distillation within the homogeneous architecture, Generative-to-Generative or Retrieval-to-Retrieval, we focus on the model-level distillation between heterogeneous architectures, especially Generative-to-Retrieval, to take advantages of each architecture.
\section{Conclusion}\label{sec:6_conclusion}
We present G2R, a novel training scheme of retrieval model for open-domain conversation by distilling the knowledge of large-scale generative models in both data-level and model-level.
G2R enables retrieval models to build a highly efficient conversation system that exhibits a substantial level of conversational ability.
We believe that our work will serve as a stepping stone for creating an efficient and real-time open-domain conversation system.
\section*{Ethical Considerations}
We train our models with the BST+ dataset, and the models we used for the pre-training (Pre-trained Bi-encoder weights from \citet{humeau2019poly}) and generating the augmented dataset (Blender 9.4B) are trained with the Pushshift Comment Dataset \citep{baumgartner2020pushshift} and the BST+ dataset.
Both the BST+ dataset and the Pushshift dataset are publicly available.
Texts included in these datasets may include potentially abusive contents and underlying biases, and these toxicities and biases could have been unintentionally encoded in our models.
Therefore, methods for reducing the toxicity of the open-domain dialogue system \citep{xu2020recipes, dinan2019build} or methods for mitigating the bias of the dialogue model \citep{liu2020mitigating, dinan2020queens} are recommended to be jointly used with our method when deploying our model in production.

Like any other open-domain conversational system, our system might provide false or misleading information.
Furthermore, our system has the potential to return a response that contains private information.
Since our model is a retrieval-based model and the pre-defined response set is fixed, an effort for filtering out the responses that potentially contain false information, private information, profanity, and inappropriate content should be preceded.

We acknowledge that it is possible to have biases in human evaluation through Amazon Mechanical Turk. To reduce potential biases, we set a maximum number of annotations per worker. We did not ask the user's identity; therefore, their personal information, including their gender, race, ethnicity, etc., is not revealed.

\bibliography{anthology,custom}
\bibliographystyle{acl_natbib}

\appendix

\end{document}


\maketitle
\section{Implementation Details}

\subsection{Baseline models}

\textbf{Blender Models.} For Blender models (\textit{Blender 90M}, \textit{Blender 2.7B}, \textit{Blender 9.4B}, \textit{Distilled Blender}), we use the pre-trained weights released from ParlAI  (Miller et al., 2017).
For a generation, we follow the decoding hyperparameters suggested from the original work (Roller et al., 2021) - using beam search with beam size 10, minimum beam length 20, and tri-gram beam blocking on context and response blocks.

\textbf{Retrieval Models.} 
We train the \textit{Bi-encoder} and the \textit{Poly-encoder} baseline model on BST+ dataset with pre-trained weights released in ParlAI (Miller et al., 2017), which is originally disclosed by Humeau et al. (2019).
Both models have a network parameter size of 256M.
We train the model with BST+ dataset, with the batch size 512 and the configuration of using other responses in batch as random negatives, initial learning rate of 1e-5, \textit{ReduceOnPleteau} learning rate schedule with decay rate 0.5 and patience 1. 
The validation Hits@1/K metric is employed as a proxy metric.
Also, we utilize Adamax optimizer (Kingma and Ba, 2015) with gradient clip value 0.1 for our experiments.
Note that most of the hyperparameters follow the default implementation of f Humeau et al. (2019) implemented in the ParlAI library.
These learning hyperparameters were also used for training other retrieval models in this paper, unless stated.

\textbf{RetNRef.}
We train the RetNRef model with a 256M Bi-encoder model architecture as a retriever and 90M Blender generative model architecture as a generator. We follow the $\alpha$-blending training scheme of  (Roller et al., 2021), using blending parameter $\alpha=0.5$. 
The model was trained with a batch size of 32 and an initial learning rate of 7e-6, with \textit{ReduceOnPleateau} learning rate scheduler with validation PPL as a proxy metric (with decay rate 0.5, patience 1).
For inference, we use the same decoding hyperparameters as in Blender generative models except for the minimum beam length constraint parameter.
We used 0 for this value since using a larger value induced a severe repeating problem in the generated response and hurt the performance of the model.

\subsection{FAISS}
FAISS (Johnson et al., 2019) is employed as an efficient MIPS library for our retrieval-based conversation pipeline.
Hierarchical Navigable Small World approximation (Malkov and Yashunin, 2018) is used for building a FAISS index, which was empirically found to be fast and accurate.
We use \textit{HNSW32\_Flat} index with \textit{efSearch} parameter 256 whlie using FAISS throughout our implementation.

\subsection{Data-level G2R}
We use the Blender 9B model (Roller et al., 2021) as our large-scale generative model $\mathcal{G}$.
Throughout our experiments, we use the BST+ training dataset as the original dialogue dataset $D$, without using the meta-information such as the persona information from ConvAI2 (Zhang et al., 2018) and the Wikipedia topic information from WoW (Dinan et al., 2018).
We use top-k sampling with $k=20$ and tri-gram beam blocking on context and response blocks. 
We sample 5 samples each from two configurations that use the beam min length hyperparameter of 10 and 20, respectively, sampling a total of 10 samples from a single context $c_i$.
We mainly used ParlAI (Miller et al., 2017) for our experiments.
For training Data-level G2R based retrieval model, we compose a mini-batch by randomly selecting 48 unique contexts and randomly selecting 10 responses connected to each context, resulting in a total of 480 (context, response) pairs in a single batch.
512 random negatives are uniformly sampled from response repository $R^\mathcal{G}$ and used as a shared random negative among the examples in the batch.
We use the Bi-encoder model trained for baseline retrieval model as initial weights and use the same learning configuration as in the baseline retrieval model except for the initial learning rate value of 5e-5.
We tested the initial learning rate value $lr \in \{1e-5, 5e-5\}$ and selected $5e-5$ since this value has shown faster convergence and higher validation Hits@1/K metric.
We trained the model until the convergence of validation Hits@1/K metric and chose the model with the best Hits@1/K metric along the training process.
Training takes about 16 to 24 hours in a single NVIDIA DGX Station A100 workstation.

\subsection{Model-level G2R}
For model-level G2R, we use hyperparameter of $\alpha=0.9$, and $T=1$.
We did not perform hyperparameter search on $T$, and tried $\alpha \in \{0.5, 0.9\}$ and selected 0.9 for $\alpha $ since 0.9 has shown higher validation Hits@1/K metric. 
We use the same training configuration as we train the model in data-level G2R.

\section{Metrics Details}

\subsection{Human evaluation}

For accurate human evaluation, we only received an annotation from turkers that satisfies the following requirements: (1) HITs approval rate greater than 95\%, (2) Location is one of Australia, Canada, New Zealand, United Kingdom, and the United States, (3) Lifetime number of HITs approved greater than 1000, following Li et al. (2018).
The instruction for the human evaluation is provided below:

\begin{quote}
Given the dialogue context, you need to rate the quality of the given response in terms of \textbf{appropriateness} and \textbf{informativeness}.

Appropriateness is a metric for evaluating whether \textbf{the given response is fluent, logical, and appropriate to its given context}. Please rate appropriateness with the range of 0 to 2, where 0 represents bad, and 2 represents excellent. Assign a lower score to the response if the response seems off (illogical, out of context, confusing).

Informativeness is a metric for evaluating whether \textbf{the given response has meaningful information relevant to its given context}. Please rate informativeness with the range of 0 to 2, where 0 represents bad, and 2 represents excellent. Please assign a higher score if the response is rich and specific to the context and a lower score if the response is bland and generic.
\end{quote}

\subsection{Measuring Latency}
We use NVIDIA DGX Station A100 for measuring the latency of the model, with Pytorch 1.7.1, Cuda 11.0, CuDNN 8.0.
We only utilize a single GPU (NVIDIA A100 GPU, 40GB Memory) for measuring the latency.
Latency is measured as the average inference time of 200 response generations after having 3 warm-up steps.

\subsection{Details for Calculating Metrics}
For calculating the MaUdE (Sinha et al., 2020) metric, we used the code provided by the authors\footnote{\url{https://github.com/facebookresearch/online_dialog_eval}}.
For calculating the Dist-2, Dist-3 metrics, and Length, we tokenized the generated response with the \texttt{casual\_tokenize} method of the \texttt{nltk} library (Loper and Bird, 2002) and calculated the metric over 200 generated responses. 

\section{CPU Latency}
\begin{table}[h]
\centering
\begin{tabular}{c|c}
\toprule
Model Name & Latency (ms) \\
\midrule
Bi-encoder & 145.5 \\
G2R-DM-LL (w/o FAISS) & 419.4 \\
G2R-DM-LL & 163.9 \\
Blender 90M & 1908.0 \\
Distilled Blender & 9295.8 \\
\bottomrule
\end{tabular}
\caption{Latency of the models measured by only using CPU.}
\label{tab:cpu_latency}
\end{table}

\begin{table*}[ht]
\centering
\footnotesize
\begin{tabular}{cccccc}
\toprule
Model Name & Valid Hits@1/K & Valid Hits@5/K & Test Hits@1/K & Test Hits@5/K \\
\midrule
Bi-encoder & 0.7469 & 0.9280 & 0.7537 & 0.9363 \\
Bi-encoder $\mathcal{R}$ in G2R-D & 0.8011 & 0.9559 & 0.8052 & 0.9570 \\
Bi-encoder $\mathcal{R}$ in G2R-DM & 0.8011 & 0.9558 & 0.8043 & 0.9601 \\
\bottomrule
\end{tabular}
\caption{Hits@N/K metrics of G2R models measured on the validation and the test split of BST+ dataset.}
\label{tab:valid_test_hits}
\end{table*}

\begin{table}[ht]
\footnotesize
\centering
\begin{tabular}{|c|l|l|l|c|c|c|}
\hline
\multicolumn{7}{|c|}{\textbf{Input Context $c_i$ }}                                                  \\ \hline
\multicolumn{7 }{|c|}{\begin{tabular}[c]{@{}c@{}}A: Hi I used to be a butcher, but I stopped.\end{tabular}} \\ \hline
\hline
\multicolumn{4}{|c|}{\textbf{Response}}        & Src.      &    $\mathcal{G}_{LL}$ & $\mathcal{G}_{MI} $ \\ \hline 
\multicolumn{4}{|p{4.3cm}|}{Why did you stop? was it too dirty? Do you have a job now?}                          &  $\mathcal{G}$ &  -1.38 & 1.16                     \\ \hline
\multicolumn{4}{|p{4.3cm}|}{That must have been an exciting job, but why did you quit? I love getting fresh meat from the butcher}         & $\mathcal{G}$                    & -1.62 & 1.37                     \\ \hline
\multicolumn{4}{|p{4.3cm}|}{Hi! You've watched the movie the chronicles of riddick?} & D & -2.55  & 0.10\\ \hline
\end{tabular}
\caption{Example responses and corresponding generator scores $\mathcal{G}_{LL}(c,r)$ (LL score) and $\mathcal{G}_{MI}(c,r)$ (MI score) for model-level G2R. \textit{Src.} column indicates the origin of each response, where $\mathcal{G}$ means it is created by the generation model $\mathcal{G}$ at data-level G2R and $D$ means the response is from original dialogue.}
\label{tab:model_level_g2r_example}
\end{table}

In Table \ref{tab:cpu_latency}, we report the latency of various models measured by using only CPU.
While retrieval models, especially \textit{Bi-encoder} and \textit{G2R-DM}, show an acceptable latency under 200ms, generation models such as \textit{Blender 90M} and \textit{Distilled Blender} exhibit inordinately high latency over 1 second.
In particular, \textit{Distilled Blender} shows the latency of 9.3 seconds.
The immensely high latency of generative models makes it extremely difficult to employ these models to build real-time conversation agents in a situation where only CPU is available for inference.

For calculating CPU latency, we utilized a Ubuntu machine with 40 Intel Xeon Silver 4210 CPU (2.20GHz) and 250GB RAM, and measured latency as the average inference time of 50 response generations after having 3 warm-up steps.

\section{Human Evaluation Details}

\begin{table*}[t]
\centering
\footnotesize
\begin{tabular}{cccc}
\toprule
\multicolumn{1}{c}{\multirow{2}{*}{Models}} & \multicolumn{3}{c}{Human Evaluation}  \\ \cmidrule(lr){2-4}
\multicolumn{1}{c}{}                        & \multicolumn{1}{c}{Sum} & \multicolumn{1}{c}{Appr.} & \multicolumn{1}{c}{Info.}   \\
\midrule
Blender 90M & 2.843$\pm$0.091  & 1.429$\pm$0.058 & 1.414$\pm$0.048 \\
Blender 2.7B & 2.983$\pm$0.091 & 1.510$\pm$0.054 & 1.473$\pm$0.053 \\
Blender 9.4B & 2.930$\pm$0.092 & 1.472$\pm$0.056 & 1.458$\pm$0.053 \\ 
Distilled Blender & 2.910$\pm$0.087 & 1.474$\pm$0.054 & 1.436$\pm$0.051 \\ 
RetNRef & 2.771$\pm$0.085 & 1.404$\pm$0.049 & 1.368$\pm$0.053    \\ 
Bi-encoder & 2.597$\pm$0.105 & 1.288$\pm$0.060 & 1.309$\pm$0.062 \\
Poly-encoder & 2.686$\pm$0.094 & 1.340$\pm$0.055 & 1.346$\pm$0.055  \\ 
Bi-encoder (w/ FAISS) & 2.596$\pm$0.096 & 1.259$\pm$0.056 & 1.337$\pm$0.055  \\
\midrule
G2R-D (w/o FAISS)  & 2.779$\pm$0.100 & 1.380$\pm$0.056 & 1.399$\pm$0.059 \\
G2R-D  & 2.759$\pm$0.109 & 1.398$\pm$0.060 & 1.361$\pm$0.064 \\
G2R-DM & 2.856$\pm$0.098 & 1.447$\pm$0.058 & 1.410$\pm$0.056 \\
G2R-DM (MI Score) & 2.806$\pm$0.098 & 1.427$\pm$0.059 & 1.380$\pm$0.056 \\
\midrule
Human Response & 2.788$\pm$0.103 & 1.418$\pm$0.058 & 1.369$\pm$0.060 \\
\bottomrule

\end{tabular}
\caption{Human evaluation results of the baseline models and our G2R models with 95\% confidence interval.} 
\label{tab:main_result_interval}
\end{table*}

\begin{table*}[t]
\centering
\footnotesize
\begin{tabular}{ccccc}
\toprule
\multicolumn{1}{c}{\multirow{2}{*}{Model A}} & \multicolumn{1}{c}{\multirow{2}{*}{Model B }} & \multicolumn{3}{c}{Human Evaluation}  \\ \cmidrule(lr){3-5}
\multicolumn{1}{c}{}                    &    & \multicolumn{1}{c}{Sum} & \multicolumn{1}{c}{Appr.} & \multicolumn{1}{c}{Info.}   \\
\midrule

G2R-D & Bi-encoder (w/ FAISS) & \textbf{0.028} & \textbf{0.001} & 0.575 \\
G2R-DM & Bi-encoder (w/ FAISS) & \textbf{<0.001} & \textbf{<0.001} & 0.073 \\
G2R-DM (MI Score) & Bi-encoder (w/ FAISS) & \textbf{0.003} & \textbf{<0.001} & 0.291 \\
G2R-D & Poly-encoder & 0.311 & 0.156 & 0.717 \\
G2R-DM & Poly-encoder & \textbf{0.014} & \textbf{0.009} & 0.112 \\
G2R-DM (MI Score) & Poly-encoder & 0.080 & \textbf{0.035} & 0.396 \\
G2R-DM & G2R-D & 0.199 & 0.255 & 0.266 \\
G2R-DM (MI Score) & G2R-D & 0.533 & 0.507 & 0.672 \\
Human Response & G2R-D & 0.715 & 0.636 & 0.860 \\
G2R-DM & G2R-DM (MI Score) & 0.481 & 0.636 & 0.459 \\
Blender 90M & Human Response & 0.427 & 0.793 & 0.252 \\
Distilled Blender & Human Response & 0.078 & 0.176 & 0.097 \\
Blender 2.7B & Human Response & \textbf{0.006} & \textbf{0.024} & \textbf{0.011} \\
Blender 9.4B & Human Response & \textbf{0.044} & 0.194 & \textbf{0.029} \\
G2R-DM & Blender 90M & 0.851 & 0.679 & 0.907 \\
Human Response & G2R-DM & 0.346 & 0.502 & 0.335 \\
Distilled Blender & G2R-DM & 0.428 & 0.505 & 0.499 \\
Blender 2.7B & G2R-DM & 0.064 & 0.116 & 0.110 \\
\bottomrule

\end{tabular}
\caption{P-value of the two-tailed t-test between two models on human evaluation results. We boldface the p-values under 0.05. Model A has a better average \textit{Sum} human evaluation score than Model B.} 
\label{tab:main_result_pvalue}
\end{table*}

We provide additional statistics about the human evaluation result, including 95\% confidence interval and p-values for two-tailed t-test between the human evaluation scores of two models in Table \ref{tab:main_result_interval} and Table \ref{tab:main_result_pvalue}, respectively.
Since the number of annotations was relatively small (200 examples) due to the cost of the human evaluation, the majority of the comparison is not statistically significant ($p<0.05$).
However, we observed that the comparison between \textit{G2R-D vs. Bi-encoder (w/ FAISS)},  \textit{G2R-DM vs. Bi-encoder (w/ FAISS)}, \textit{G2R-DM (MI Score) vs. Bi-encoder (w/ FAISS)} and \textit{G2R-DM vs. Poly-encoder} shows a statistically significant difference in terms of Sum of human evaluation score and the Appropriateness human evaluation score, proving that our G2R methods improve the performance of the retrieval model.
Also, note that the trend of the \textit{Sum} human evaluation score within 90M, 2.7B, and 9.4B Blender models is similar to the trend of ACUTE-Eval Engagingness evaluation result reported in the original paper (Roller et al.,
2021), which adds more reliability to our human evaluation result.

\section{Dataset Details}
BST+ dataset is a concatenation of four English dialogue dataset (Blended Skill Talk (Smith et al.,
2020), ConvAI2 (Zhang et al., 2018), Empathetic Dialogues (Rashkin et al., 2019) and Wizard of Wikipedia (Dinan et al., 2018)).
We use the Blender 9.4B model to augment this dataset as described in the \textit{Data-level G2R} section, and the augmented dataset consists of total 3,070,033 context-response pairs on 274,233 unique contexts.
As described in the Experiments section, we release the augmented BST+ dataset in \url{https://github.com/hyperconnect/g2r}.

\section{Validation and Test Hits@1/K metrics}
For reference, we report the Hits@1/K and the Hits@5/K metrics of our retrieval models measured on the validation and the test split of BST+ in Table \ref{tab:valid_test_hits}. 

\section{Details for Model-level G2R Ablation Study}
Here, we provide additional details for calculating the MI score in the ablation study for model-level G2R.
MI score is calculated with the \textit{MMI-bidi} equation in the original paper (Li et al., 2016), but  additionally normalized by the length of response in the same way LL score is normalized:
\begin{equation}
    \mathcal{G}_{MI}(c, r) = (\log P_{\mathcal{G}}(r|c) - log P_{\mathcal{G}}(r)) / |r|.
    \label{eq:mutual_information_score}
\end{equation}
Since calculating the unconditional language probability term $P_\mathcal{G}(r)$ in Equation \ref{eq:mutual_information_score} is intractable, we approximate this term by taking the average of the likelihood values of $r$ given dummy input contexts, including \textit{"."}, \textit{"<PAD>"} and \textit{"<UNK>"}.
This trick enables us to avoid undesirable alternative options for calculating $P_\mathcal{G}(r)$ with high computational burden, such as training a separate unconditional language model or calculating an intractable marginal probability $\sum_c P_\mathcal{G}(r|c) P(c)$.

\section{Data-level and Model-level G2R Examples}
Table \ref{tab:model_level_g2r_example} shows the example responses generated by the data-level G2R, and LL and MI score calculated for each response.
Data-level G2R is able to generate high-quality responses that are appropriately related to the input context.
Model-level G2R helps the retrieval model to distinguish between low-quality and high-quality responses since both LL and MI scores assign a low value for an inappropriate response.
Comparing both scores in this example, the MI score assigns higher values to a relatively more specific response compared to the LL score.
In comparison, the LL score assigns higher values to a relatively more general response that can also be appropriately utilized for a different context.

\section{Additional Response Examples}

\begin{table*}[ht]
\centering
\footnotesize

\begin{tabular}{c|l|l|l|c}
\toprule
\multicolumn{5}{c}{\textbf{Input Context}}                                                  \\ \midrule
\multicolumn{5}{l}{\begin{tabular}[l]{@{}l@{}} A: I've been doing it as a hobby for about five years. I'm still not great but I keep improving. \\ B: That is nice!!! I am usually very punctual. I make it early to every class. \\ A: That's a great quality to have. It will help you move up later in life. When you start working that will \\ make a great impact on your boss. \end{tabular}} \\ 
\midrule
\multicolumn{1}{c|}{\textbf{Model Name}}                 &  \multicolumn{4}{|c}{\textbf{Response}}                 \\ \midrule
Bi-encoder+FAISS & \multicolumn{4}{|l}{B: Definitely. Hopefully lead to a promotion for you soon!}  \\
Blender 90M & \multicolumn{4}{|l}{B: That is very true. I will have to keep that in mind when I start my new job.}  \\
Distilled Blender & \multicolumn{4}{|l}{B: I hope so. I need to get my grades up so I can get a good job.}  \\
Blender 2.7B & \multicolumn{4}{|l}{B: I hope so. I want to be the best at what I do. What do you do for a living?} \\
Blender 9.4B & \multicolumn{4}{|l}{B: I hope so. I want to open my own studio one day. What do you do for a living?} \\
\midrule
G2R-D & \multicolumn{4}{|l}{B: Yes it should. I can't wait to get started. I start in august.} \\
G2R-DM & \multicolumn{4}{|l}{B: I think productivity is important. I am glad I have the opportunity to work towards something.} \\
G2R-DM (MI Score) & \multicolumn{4}{|l}{B: That is great advice! I like to work hard and not let things bother me.} \\
\bottomrule
\end{tabular}

\vspace{1em}
\begin{tabular}{c|l|l|l|c}
\toprule
\multicolumn{5}{c}{\textbf{Input Context}}                                                  \\ \midrule
\multicolumn{5}{l}{\begin{tabular}[l]{@{}l@{}} (Further dialogue history omitted for space) \\ A: it is ok I am not in a hurry to be honest \\ B: were you in the military \\ A:never been in the military how about you ? \end{tabular}} \\ 
\midrule
\multicolumn{1}{c|}{\textbf{Model Name}}                 &  \multicolumn{4}{|c}{\textbf{Response}}                 \\ \midrule
Bi-encoder+FAISS & \multicolumn{4}{|l}{B: Not in the army, but I was in the fbi, g13 classified}  \\
Blender 90M & \multicolumn{4}{|l}{B: I was in the army for a few years before I joined the air force. It is a lot of work}  \\
Distilled Blender & \multicolumn{4}{|l}{B: I was in the navy for a few years. What do you do for a living?}  \\
Blender 2.7B & \multicolumn{4}{|l}{B: I was in the army for 8 years. I was a medic. What did you do?} \\
Blender 9.4B & \multicolumn{4}{|l}{B: Yes I was in the marine corps for 6 years and now I am in the reserves} \\
\midrule
G2R-D & \multicolumn{4}{|l}{B: No I never did but I was thinking about it. Do you like the army? Do u have any pets?} \\
G2R-DM & \multicolumn{4}{|l}{B: I was in it for 10 years and 2 deployments to iraq and afghanistan} \\
G2R-DM (MI Score) & \multicolumn{4}{|l}{B: I was in a combat zone but was never deployed} \\
\bottomrule
\end{tabular}

\vspace{1em}
\begin{tabular}{c|l|l|l|c}
\toprule
\multicolumn{5}{c}{\textbf{Input Context}}                                                  \\ \midrule
\multicolumn{5}{l}{\begin{tabular}[l]{@{}l@{}} A: I have this cat that loves to cause issues \\ B: What sort of issues? \\ A: Like last night. She decided that she wanted to jump on the table \end{tabular}} \\ 
\midrule
\multicolumn{1}{c|}{\textbf{Model Name}}                 &  \multicolumn{4}{|c}{\textbf{Response}}                 \\ \midrule
Bi-encoder+FAISS & \multicolumn{4}{|l}{B: Oh no. Pets are good for anger.}  \\
Blender 90M & \multicolumn{4}{|l}{B: Oh no, that's not good. I'm glad you're able to take care of her.}  \\
Distilled Blender & \multicolumn{4}{|l}{B: Oh no! I hope she doesn't do that again. Do you have any other pets?}  \\
Blender 2.7B & \multicolumn{4}{|l}{B: Oh no! I hope she doesn't do that too often. Cats can be pretty destructive.} \\
Blender 9.4B & \multicolumn{4}{|l}{B: Oh no! I hope she didn't do too much damage. Cats can be so destructive.} \\
\midrule
G2R-D & \multicolumn{4}{|p{12cm}}{B: That's hilarious. Is she a weird cat? I have a cat that will do the same thing if he's not feeling well or doesn't get what he feels like he should.} \\
G2R-DM & \multicolumn{4}{|l}{B: Oh yea, cats tend to do that!} \\
G2R-DM (MI Score) & \multicolumn{4}{|l}{B: Oh no! That doesn't really sound like a good time. What is her name? Mine is named einstein.} \\
\bottomrule
\end{tabular}

\caption{Additional example responses in the BST+ test set example.}
\label{tab:qualitative_results_suppl}
\end{table*}
\begin{table*}[ht]
\centering
\footnotesize
\begin{tabular}{c|l|l|l|c}
\toprule
\multicolumn{5}{c}{\textbf{Input Context}}                                                  \\ \midrule
\multicolumn{5}{l}{\begin{tabular}[l]{@{}l@{}} B: Hey, jefferson here, I love documentaries \\
A: I love dogs her name is barnaby, a german shepherd \\
B: Oh that's so cool. I have a dog with my husband in pennsylvania. \\
A: Nice to meet you, I work in the usps \\
B: I order a lot of books online. I read one every week. \\ 
A: How big are the dogs? Are they grey color? That's my color \\
B: I would not know, they are your dogs haha. \\
A: Do you order from amazon? My dad is a nascar racer \\
B: I do order from amazon, I'm a pa right now. \\ 
A: Are there scorpions in your area? I hate them \\
B: I sure hope not, they are scary! \\
 A: Amazon is a great place to order books \\ B: Oh yeah that is a fact. \end{tabular}} \\ 
\midrule
\multicolumn{1}{c|}{\textbf{Model Name}}                 &  \multicolumn{4}{|c}{\textbf{Response}}                 \\ \midrule
Bi-encoder (w/ FAISS) & \multicolumn{4}{|l}{A: Comics is at the top of the list in interest}  \\
Blender 90M & \multicolumn{4}{|l}{A: Do you have any other hobbies that you like to do on the weekends? I like to read}  \\
Distilled Blender & \multicolumn{4}{|l}{A: What else do you like to do in your spare time? I like to go to the beach}  \\
Blender 2.7B & \multicolumn{4}{|l}{A: What kind of books do you buy on amazon? I like to read science fiction and fantasy} \\
Blender 9.4B & \multicolumn{4}{|l}{A: What kind of books do you like to read? I like romance novels and mystery novels} \\
\midrule
G2R-D & \multicolumn{4}{|l}{A: Do you ever order grocerys? I love amazon's selection of books and videos} \\
G2R-DM & \multicolumn{4}{|l}{A: What is your favourite book? Mine is "the cat in the hat" by dr seuss} \\
\bottomrule
\end{tabular}

\caption{Example responses with full dialogue history in the case study example.}
\vspace*{-1em}
\label{tab:qualitative_results_full}
\end{table*}

We provide additional examples of responses generated by the baseline models and our G2R models in Table \ref{tab:qualitative_results_suppl}.
Also, the full dialogue context for the case study example in the main paper is provided in \ref{tab:qualitative_results_full}.

